\pdfoutput=1

\documentclass[11pt,a4paper]{article}
\usepackage{authblk}
\usepackage[hyperref]{emnlp2020}
\usepackage{times}
\usepackage{latexsym}

\usepackage{microtype}

\usepackage{microtype}
\usepackage{comment}
\usepackage{placeins}

\usepackage{multicol}

\aclfinalcopy %

\newif\ifwithappendix\withappendixtrue

\newif\ifappendixshown
\appendixshowntrue

\usepackage[T1]{fontenc}
\usepackage[utf8]{inputenc}
\usepackage[textsize=tiny]{todonotes}
\usepackage{graphicx}
\usepackage{booktabs}
\usepackage{latexsym}

\usepackage{footnote}
\makesavenoteenv{tabular}
\makesavenoteenv{table}
\makesavenoteenv{table*}

\newcommand\minput[1]{%
  \input{#1}%
  \ifhmode\ifnum\lastnodetype=11 \unskip\fi\fi}

\usepackage{hyperref}
\usepackage{xstring}

\newcommand{\noqa}[1]{}
\newcommand{\noqall}[1]{}

\newcommand{\expnumber}[2]{{#1}\mathrm{e}{#2}}

\usepackage{multirow,tabularx}
\usepackage{makecell,bigdelim}

\usepackage{amsmath,amsfonts,amssymb}
\usepackage{recycle}

\title{From Dataset Recycling to Multi-Property Extraction and Beyond}

\author[1,2]{\bf Tomasz Dwojak}
\author[1,3]{\bf Michał Pietruszka}
\author[1,4]{\bf Łukasz Borchmann}
\author[1,3]{\\ \bf Jakub Chłędowski}
\author[1,2]{\bf Filip Graliński}
\affil[1]{Applica.ai}
\affil[2]{Faculty of Mathematics and Computer Science, Adam Mickiewicz University in Poznan}
\affil[3]{Faculty of Mathematics and Computer Science, Jagiellonian University}
\affil[4]{Institute of Computing Science, Poznan University of Technology}
\affil[ ]{ }
\affil[ ]{\tt {tomasz.dwojak@applica.ai}}

\begin{document}
\maketitle
\begin{abstract}
This paper investigates various Transformer architectures on the WikiReading Information Extraction and Machine Reading Comprehension dataset.
The proposed dual-source model outperforms the current state-of-the-art by a~large margin. 
Next, we introduce WikiReading Recycled---a~newly developed public dataset, and the task of multiple-property extraction.
It uses the same data as WikiReading but does not inherit its predecessor's identified disadvantages.
In addition, we provide a~human-annotated test set with diagnostic subsets for a~detailed analysis of model performance.

\end{abstract}

\def \WR {WikiReading}%
\def \WRR {WikiReading Recycled}%
\def \MF {Mean-$F_1$}
\def \MFShort {M-$F_1$}
\def \MMF {Mean-Multi-Property-$F_1$}

\def \MMFShort {MMP-$F_1$}
\def \MPE {Multi-Property Extraction}

\section{Introduction}

The emergence of attention-based models has revolutionized Natural Language Processing~\cite{DBLP:journals/corr/abs-1708-02709}. Pretraining these models on large corpora like BookCorpus~\cite{DBLP:journals/corr/ZhuKZSUTF15} has been shown to yield a~reliable and robust base for downstream tasks. These include Natural Language Inference~\cite{snli:emnlp2015}, Question Answering~\cite{rajpurkar2016squad}, Named Entity Recognition~\cite{Yadav2018ASO,Goyal2018RecentNE,Li2018ASO}, and Property Extraction~\cite{hewlett-etal-2016-wikireading}.

The creation of large supervised datasets often comes with trade-offs, such as one between the quality and quantity of data. For instance, the \WR\ dataset~\cite{hewlett-etal-2016-wikireading} has been created in such a~way that WikiData annotations were treated as the expected answers for related Wikipedia articles. However, the above datasets were created separately, and the information content of both sources overlaps only partially. Hence, the resulting dataset may contain noise.

The best models can achieve results better than the human baseline across many NLP datasets such as MSCQAs~\cite{Wang_2018}, STS-B, QNLI~\cite{2019t5}, CoLA or MRPC~\cite{Wang2020StructBERTIL}. 
However, as a~consequence of different kinds of noise in the data, they rarely maximize the score metric~\cite{glsceil}.
While current work in NLP is focused on preparing new datasets, we regard recycling the current ones as equally important as creating a~new one. Thus, after outperforming previous state-of-the-art on \WR{}, we investigated the dataset's weaknesses and created an entirely new, more challenging \MPE{} task with improved data splits and a~reliable, human-annotated test set.

\paragraph{Contribution.} The specific contributions of this work are the following.
We analyzed the \WR\ dataset and pointed out its weaknesses. %
We introduced a~\MPE{} task by creating a~new dataset: \WRR{}. Our dataset contains a~human-annotated test set, with multiple subsets aimed to benchmark qualities such as generalization on unseen properties.
We introduced a~\MMF\ score suited for the new \MPE{} task.
We evaluated previously used architectures on both datasets. Furthermore, we showed that pretrained transformer models (Dual-Source RoBERTa and T5) beat all other baselines.
The new dataset and all the models mentioned in the present paper were made publicly available on GitHub.\footnote{\url{https://github.com/applicaai/multi-property-extraction}}

\section{Related Work}

\begin{table*}
\renewcommand{\arraystretch}{0.9}
\centering
\begin{tabular}{llll}
\toprule
Dataset & Task & Input & Output \\ \midrule
SNLI & Natural Language Inference & two sentences & relation between the sentences %
\\
SQUAD & Question Answering & article, question & answer to the question \\
WiNER & Named Entity Recognition & article & annotated named entities \\
WR & Property Extraction & article, property & value of the property  \\
WRR (ours) & Multi-Property Extraction & article, properties & values of the properties \\ \bottomrule
\end{tabular}
\caption{Comparison of NLP tasks on text comprehension and information extraction. More differences between WR and WRR were outlined in Table~\ref{tab:wrwrrmango}.}
\label{tab:tasks}
\end{table*}

Early work in relation extraction revolves around problems crafted using distant supervision methods, which are ~semi-supervised methods that automatically label pools of unlabeled data ~\cite{Craven:1999:CBK:645634.663209}.
In contrast, many QA datasets were created through crowd-sourcing, where annotators were asked to formulate questions with answers that require knowledge retrieval and information synthesis. One of the most popular QA datasets is Wikipedia-based SQUAD, where an instance consists of a~human-formulated question, and an encyclopedic reading passage used to base the answer on~\cite{rajpurkar-etal-2018-know}.
Another crowd-sourced dataset that profoundly influenced Natural Language Inference research is SNLI~\cite{snli:emnlp2015}---a~three-way semantics-based classification of a~relation between two different sentences. 

Both SQUAD and SNLI are large-scale Machine Reading Comprehension (MRC) tasks, but they cannot be treated as Property Extraction as defined in Section \ref{s:property_extraction}; hence they are not considered in this paper.
Similarly, some MRC problems framed in TREC tracks, such as Conversational Assistance or Question Answering, are beyond the scope of this paper~\cite{dalton2020trec, TREC_2007_Question_Answering}.

\citet{hewlett-etal-2016-wikireading} proposed the \WR\ dataset that consists of a~Wikipedia article and related WikiData statement. %
No additional annotation work was performed, yet the resulting dataset was of presumably high reliability. Nevertheless, we consider an additional human annotation to be desired (Section \ref{sec:annotations}). Alongside the dataset, a~property extraction task was introduced. The idea behind it is to read an article given a~property name and to infer the associated value from the article. The property extraction paradigm is described in detail in Section~\ref{s:property_extraction}, whereas a~brief comparison to related datasets is presented in Table~\ref{tab:tasks}.

Initially, the best-performing model used placeholders to allow rewriting out-of-vocabulary words to the output.
Next, \citet{choi-etal-2017-coarse} presented a~reinforcement learning approach that improved results on a~challenging subset of the 10\% longest articles. %
This framework was extended by~\citet{MSCQA_System} with a~self-correcting action that removes the inaccurate answer from the answer generation module and continues to read.

\citet{hewlett-etal-2017-accurate} hold the state-of-the-art on \WR\ with their proposition of SWEAR that attends over a~sliding window's representations to reduce documents to one vector from which another GRU network generates the answer~\cite{chung2014empirical}. Additionally, they evaluated a~strong semi-supervised solution on a~randomly sampled~1\% subset of \WR. 

To the best of our knowledge, no authors validated Transformer-based models on \WR{} and pretrained encoders.

\section{Property Extraction}\label{s:property_extraction}

Let a~\textit{property} denote any query for which a~system is expected to return an answer from given text.
Examples include \textit{country of citizenship} for a~biography provided as an input text, or \textit{architect name} for an article regarding the opening of a~new building. Contrary to QA problems, a~query is not formulated as a~question in natural language but rather as a~phrase or keyword.
We use the term \textit{value} when referring to a~valid answer for the stated query. Some properties have multiple valid answers; thus, multiple values are expected. Examine the case of Johann Sebastian Bach's biography for which property \textit{sister} has eight values.
We will refer to any task consisting of a~tuple (properties, text) for which values are to be provided as a~property extraction task.

\begin{table}
\renewcommand{\arraystretch}{0.9}
\centering
\begin{tabular}{lrrr}
\toprule
Data split & Size & In train & \% \\ \midrule
Validation set &  1,452,591 & 1,374,820 & 94.65 \\
Test set &  821,409 & 780,639 & 95.04 \\ \bottomrule
\end{tabular}
\caption{The size of \WR\ splits (\emph{Size}) and number of  articles leaked from the train set as an absolute value or percentage.}

\label{tab:wr-overlap}
\end{table}

The biggest publicly available dataset for property extraction is \WR{}~\cite{hewlett-etal-2016-wikireading}. The dataset combines articles from Wikipedia with Wikidata information. %
The dataset is of great value; however, several flaws can be identified. First, more than 95\% of articles in the test set appeared in the train set (Table~\ref{tab:wr-overlap}). Second, the unjustifiably large size of the test set is a~substantial obstacle for running experiments. For instance, it takes 50 hours to process the test set using a~Transformer model such as T5$_\textsc{SMALL}$ on a~single NVidia V100 GPU. Finally, \WR{} assumes that every value in the test set can be determined on the basis of a~given article. As shown later, this is not the case for 28\% of values.

\subsection{Towards Multi-Property Extraction}

In the \MPE{}~(MPE) scenario we propose, the system is expected to return values for multiple properties at once. Hence, can be considered a~generalization of a~single-property extraction task as it can be easily formulated as such. 
Thus, MPE is reverse-compatible with the single-property extraction, and it is still possible to evaluate models trained in the single-property setting.

Many arguments can be considered in favor of framing the problem as MPE. In a~typical business scenario, multiple properties are expected to be extracted from a~given document. The bulk inference requires a~lower computational budget by a~factor proportional to the mean number of properties per article, which makes MPE preferable.
Moreover, one can expect that systems trained in such a~way will manifest emergent properties resulting from the interaction between properties themselves. Consider the set of property-value pairs:
\begin{quote}
    \small
    date of birth: 1915-01-12, date of death: 1979-05-02, place of birth: Saint Petersburg
\end{quote}
already predicted by an autoregressive model. It is in principle possible to answer:
\begin{quote}
    \small
    country of citizenship: Russian Empire,  country of citizenship: Soviet Union 
\end{quote}
using the earlier predicted pairs only. This phenomenon emerges if the model (or person) learned the relationships between years, administrative boundaries of the city, and the transformation of the Russian Empire into a~communist state that occurred in the meantime. Although no such reasoning is required and the problem can be solved by memorizing related co-occurrence patterns, we intend to achieve the mentioned emergent properties.

\section{WikiReading Recycled: Novel Dataset for Multi-Property Extraction}

\begin{table}
\renewcommand{\arraystretch}{0.9}
\centering
\begin{tabular}{lcc}
\toprule
Feature & WR & WRR \\ \midrule
Base unit & property & article \\
Examples & $18.6$M  & $4.1$M \\
Properties/example &  $1$ & $4.5$ \\
\midrule
Metric & \MFShort\ & \MMFShort \\
Human-annotated test & $-$ & $+$ \\
\midrule
Dataset split & random & controlled \\
Unseen in evaluation  & $-$ & $+$ \\
Article appears in & few splits & one split \\
\bottomrule
\end{tabular}
\caption{Selected differences between WR and WRR. Both metrics are described in Section~\ref{s:evaluation}.}
\label{tab:wrwrrmango}
\end{table}
The comparison to existing datasets and shared tasks is briefly presented in Table~\ref{tab:tasks}, whereas Table~\ref{tab:wrwrrmango} focuses on selected differences between \WRR{} and \WR{}.

\subsection{Desiderata \label{sec:desiderata}}
Our set of desiderata is based on the following intentions. We wished to introduce the problem of \MPE{} to evaluate systems that extract any number of given properties at once from the same source text. Our second objective was to ensure that an article may appear in precisely one data split. The third core intention was to introduce an article-centered data objective instead of a~property-centric one. Note that an instance of data should be an article with multiple properties. The fourth objective was to ensure that all properties in the test set can be extracted or inferred. The fifth was to keep the validation and test sets within a~reasonable size. Moreover, we aim to provide a~test set of the highest quality, lacking noise that could arise from automatic processing.
Finally, we intended to benchmark the model generalization abilities -- the test set contains properties not seen during training, posing a~challenge for current state-of-the-art systems.

\subsection{Data Collection and Split} \label{sec:Description of WRR}

The \WRR{} and \WR{} are based on the same data, yet differ in how they are arranged. %
Instances from the original \WR{} dataset were merged to produce over 4M samples in the MPE paradigm. Instead of performing a~random split, we carefully divide the data assuming that 20\% of properties should appear solely in the test set (more precisely, not seen before in train and validation sets).
Around one thousand articles containing properties not seen in the remaining subsets were drafted to achieve the mentioned objective. Similarly, properties unique for the validation set were introduced to enable approximation of the~test set performance without disclosing particular labels.
Additionally, test and validation sets share 10\% of the properties that do not appear in the train set, increasing the size of these subsets by 2,000 articles each. Another 2,000 articles containing the same properties as the train set were added to each of the validation and test sets. All the remaining articles were used to produce the training set.

To sum up, we achieved a~design where as much as 50\% of the properties cannot be seen in the training split, while the remaining 50\% of the properties can appear in any split. We chose these properties carefully so that the size of the test and validation sets does not exceed 5,000 articles.

\begin{table}
\renewcommand{\arraystretch}{0.9}
\centering
\begin{tabular}{lrrrr}
\toprule
Subset & Dev & Test-A & Test-B \\ \midrule
rare & $4.40$ & $5.12$ & $3.16$ \\
unseen & $5.53$ & $5.34$ & $2.05$ \\
categorical & $46.63$ & $44.49$ & $66.51$ \\
relational & $53.36$ & $55.50$ & $33.49$ \\
exact match & $20.20$ & $20.16$ & $33.67$ \\
long articles  & $50.39$ & $56.15$ & $30.45$ \\

\bottomrule
\end{tabular}
\caption{An average per-article size of the corresponding subsets as a~percent of a~total number of properties.\label{tab:subsets}}
\end{table}

\subsection{Human Annotation\label{sec:annotations}}
The quality of test sets plays a~pivotal role in reasoning about a~system's performance. Therefore, a~group of annotators went through the instances of the test set and assessed whether the value either appeared in the article or can be inferred from it. To make further analysis possible, we provide both datasets, before (test-A) and after (test-B) annotation.

The annotation process was non-trivial due to vagueness of the inferability definition, and the scientific character of the considered text. It was required to understand advanced encyclopedic articles e.g., about chemistry, biology, or astronomy, to answer domain-specific properties (scientific classifications or biological taxonomy), which are only possible with deep knowledge about the world and with the ability to learn during the process. Moreover, linguistic skills were required to transliterate and transcribe first and last names. Note that we consider the value which appears in a~different writing script as inferable. Due to the stated issues, we decided to rely on highly trained linguists as annotators.

The process was supported by several heuristics. In particular, the approximate string matching was used to highlight fragments of presumably high importance. 
Nevertheless, it took seven linguists more than 100 hours in total to complete. On average, two minutes and thirty second were required to verify data assigned to one Wikipedia article.

The relevance of annotation mentioned above can be demonstrated by the fact that 28\% of the property-value pairs were marked as unanswerable and removed.
As it will be shown later, the \MMF{} on a~pre-verified test-A was approximately $20$ points lower, and 8\% of articles were removed entirely from the test-B during the annotation process.

\subsection{Diagnostic Subsets\label{sec:diagnostics}}
We determined auxiliary validation subsets with specific qualities, not only to help improve data analysis but also to provide additional information at different stages of development of a~system.
The qualities we measure and the definition is provided below.

\paragraph{Rare, unseen.} \textit{Rare} and \textit{unseen} properties were distinguished depending on their frequency. The number of occurrences in the train set was below a~threshold of $4000$ for each in \textit{rare} and was precisely $0$ for the \textit{unseen} category.

\paragraph{Categorical, relational.} We denote a~property as \textit{categorical} if its value set contains a~limited number of values; otherwise, it is \textit{relational}. We apply normalized entropy with a~threshold of $0.7$ to obtain properties that belong to the \textit{categorical} subset.
For instance, the \textit{continent} property occurs 20060 times, but with 13 possible values, its normalized entropy equals $0.43$; hence it is marked as \textit{categorical}. This splitting method is not ideal, but we wanted to use the same method as in~\cite{hewlett-etal-2016-wikireading}. For example, if the distribution of continents was uniform, the property would have been classified as relational. However, in practice, it almost never happens.

\paragraph{Exact match.} The \textit{exact match} category applies to cases where expected value is mentioned directly in the source text.

\paragraph{Long articles.} Instances with articles longer than $695$ words (threshold qualifying to the top $15\%$ longest articles in the train set) constitute the \textit{long articles} diagnostic set.

\vspace{4mm}
Characteristics of different systems can be compared qualitatively by evaluating on these subsets. For instance, the \textit{long articles} subset is challenging for systems that consume truncated inputs. \textit{Unseen} is precisely constructed to assess systems' ability to extract previously not seen properties. On the other hand, \textit{rare} can be viewed as an approximation of the system's performance on a~lower-resource downstream extraction task. The \textit{categorical} subset is useful in assessing approaches featuring a~classifier, whereas it is suboptimal to use such systems for \textit{relational} due to richer output space. Similarly, the \textit{exact match} can be approached with sequence tagging solutions. The share of each diagnostic subset is presented in Table~\ref{tab:subsets}.

\section{Model Architectures\label{sec:transformer}}

\begin{table*}
\renewcommand{\arraystretch}{0.9}
\centering
\begin{tabular}{lccccc}
\toprule
& Basic seq2seq & \makecell{Vanilla \\ Transformer} & \makecell{Vanilla \\ Dual-Source } & \makecell{Dual-Source \\ RoBERTa} & T5 \\ \midrule
Numer of inputs & $1$ & $1$ & $2$ & $2$ & $1$ \\
Pretrained encoder & $-$ & $-$ & $-$ & $+$ & $+$ \\ 
Pretrained decoder & $-$ & $-$ & $-$ & $-$ & $+$ \\
Number of parameters & $32$M & $46$M & $25$M & $234$M & $60$M \\
\bottomrule
\end{tabular}
\caption{Comparison of evaluated models. The T5 model can be considered as a~pretrained equivalent of Vanilla Transformer, and our RoBERTa-based model can be viewed as a~partially-pretrained Vanilla Dual-Source Transformer. Basic seq2seq is an RNN counterpart of both T5 and Vanilla Transformer.
}
\label{tab:model-comparison}
\end{table*}

We evaluate different model architectures on the \WRR{} dataset. We re-implemented the previously best performing \WR{} model, finetuned pretrained Transformer models, and applied a~dual-source model. Their competitiveness can be demonstrated by the fact that we were able to outperform the previous state-of-the-art on the \WR{} by a~far margin. 

\paragraph{Basic seq2seq.} A~straightforward approach to single-property extraction is to use an~LSTM sequence-to-sequence model where the input consists of a~property name concatenated with the considered input text. To compare with the previous results, we reproduced the basic sequence-to-sequence model proposed by~\citet{hewlett-etal-2016-wikireading}.

\paragraph{Vanilla Transformer.} A~more up-to-date solution is to use the Transformer architecture~\cite{NIPS2017_7181} instead of an RNN, and a~subword tokenization method, such as unigram LM tokenization~\cite{kudo-2018-subword}. We use the term \textit{vanilla} to denote a~model trained from scratch.

\begin{figure}
    \centering
    \includegraphics[width=0.9\linewidth]{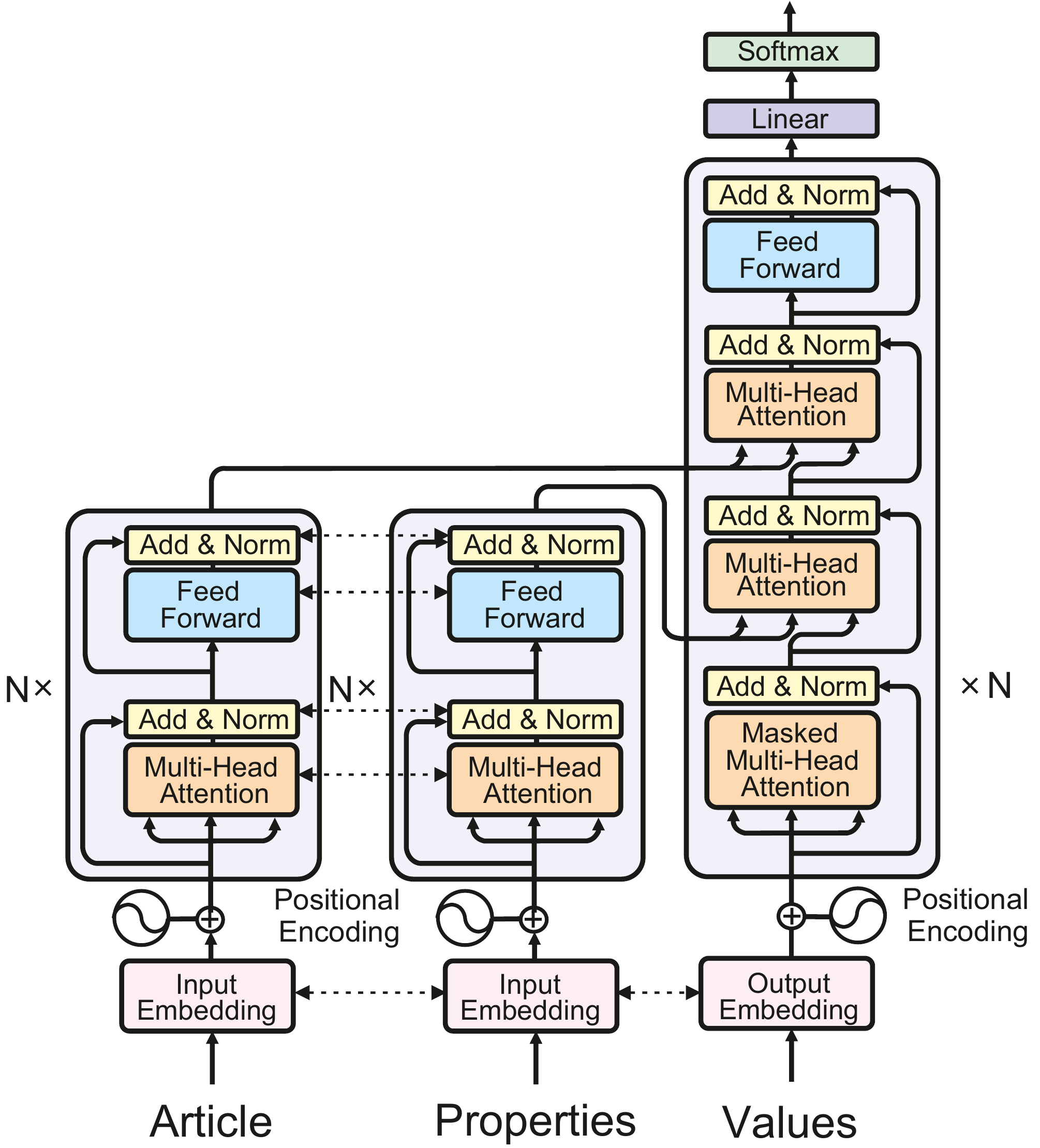}
    \caption{The architecture of Dual-Source Transformer as proposed by~\citet{junczys-dowmunt-grundkiewicz-2018-ms} for Automatic Post-Editing. In the case of \WRR{} and \WR{}, the encoder transforms an article and the corresponding properties separately.
    }\label{fig:architecture}
\end{figure}

\paragraph{Vanilla Dual-Source Transformer.}  The Transformer architecture was extended to support two inputs and successfully applied in Automatic Post-Editing~\cite{junczys-dowmunt-grundkiewicz-2018-ms}. We propose to reuse this Dual-Source Transformer architecture in~the property extraction tasks. The architecture consists of two encoders that share parameters and a~single decoder. Moreover, the encoders and decoder share embeddings and vocabulary. In our approach, the first encoder is fed with the text of an article, and the second one takes the names of properties (Figure~\ref{fig:architecture}). The model is trained to generate a~sequence of pairs: (\textit{property}, \textit{value}) separated with a~special symbol.

\paragraph{Dual-Source RoBERTa.} Recent research shows that pretrained language models can improve performance on downstream tasks~\cite{Radford2018ImprovingLU}. Therefore, we experimented with the pretrained RoBERTa language model as an encoder. RoBERTa models were developed as a~hyper-optimized version of BERT with a~byte-level BPE and a~considerably larger dictionary~\cite{Liu2019RoBERTaAR, DBLP:journals/corr/abs-1810-04805}. All the model parameters, including the RoBERTa weights, were further optimized on the \WRR{} task. 

\paragraph{T5.} Recently proposed T5 model~\cite{2019t5} is a~Transformer model pretrained on a~cleaned version of CommonCrawl. T5 is famous for achieving excellent performance on the SuperGLUE benchmark~\cite{DBLP:journals/corr/abs-1905-00537}.

To create a~model input, we concatenate a~property name and an article. In the case of MPE, we reduce the dataset to the single property setting, as used by the T5 model's authors.

\section{Evaluation\label{s:evaluation}}
In this section, we describe the evaluation of previously proposed architectures on both \WR{} and \WRR{} datasets. We would like to highlight that the results are not comparable between the two datasets, as they are based on different train/validation/test splits.

\subsection{Metrics}
The performance of systems is evaluated using the F1 metric, adapted for the \WRR\ format. For \WR{}, \MF\ follows the originally proposed micro-averaged metric and assesses F1 scores for each property instance, averaged over the whole test set. 

Let $E$ denote a~set of expected property-value pairs and $O$ model-generated property-value pairs. Assuming $\left| \cdot \right|$ stands for set cardinality, precision and recall can be formulated as follows:
\begin{equation*}
P(E, O) = \frac{\left| E \cap O \right|}{\left| O \right|
}
,\text{  }
R(E, O) = \frac{\left| E \cap O \right|}{\left| E \right|
}
\end{equation*}
Then $F_1$ is computed as a~harmonic mean:
\begin{equation*}
F_1(E, O) = 2 \cdot \frac{ P(E, O) \cdot R(E, O) }{P(E, O) + R(E, O)}
\end{equation*}

Given a~sequence $\mathcal{E} = \{E_1, E_2, .. , E_n\}$ of expected answers for $n$ test instances, and associated sequence of predictions $\mathcal{O} = \{O_1, O_2, .. , O_n\}$, we calculate \MF{} as:
\begin{equation*}
\text{Mean-}F_1(\mathcal{E}, \mathcal{O}) = \frac{1}{n} \cdot \sum_{i\in [1, n]}{F_1(E_i, O_i)}
\end{equation*}
In \WRR{}, we adjust the metric to handle many properties in a~single test instance. To do that, the $E_i$ and $O_i$ sets contain values from many properties at once and $n$ is equal to the number of articles. Note that in the case of the \MFShort{} properties are considered as instances.  We call our article-centric metric \MMF{} or in short \MMFShort.

\subsection{Training Details\label{sec:baseline}}

Since the basic seq2seq model description missed some essential details, they had to be assumed before model training. For example, we supposed that the model consisted of unidirectional LSTMs %
and truecasing was applied to the output. %
The rest of the parameters followed the description provided by the authors.

An extensive hyperparameter search was conducted for both Dual-Source Transformers on the \WRR\ task. In the case of the Dual-Source Transformer evaluated on \WR\, we restricted ourselves to hyperparameters following the default values specified in the Marian NMT Toolkit~\cite{mariannmt}. The only difference was the reduction of encoder and decoder depths to 4.

For the Vanilla Dual-Source Transformer evaluation, both \WR\ and \WRR\ datasets were processed with a~SentencePiece model~\cite{kudo-2018-subword} trained on a~concatenated corpus of inputs and outputs with a~vocabulary size of 32,000. %
Dual-Source RoBERTa model is initialized with RoBERTa$_\textsc{BASE}$ (consisting of 12 encoder layers and a~dictionary of 50,000 subword units).

In the case of the T5 model, we keep hyperparameters as close as possible to those used during pretraining. The training continues with restored AdaFactor parameters. We finetuned the \textit{small} version of the model in a~supervised-only manner.

We truncate the input to the first 512 tokens for all our models.

\paragraph{Hyperparameter Optimization.}  Hyperparameters for \WRR\ were optimized using the Tree-structured Parzen Estimator algorithm~\cite{NIPS2011_4443} with additional heuristics and Gaussian priors resulting from the default settings proposed for this sampler in the Optuna framework~\cite{10.1145/3292500.3330701}.
An evaluation was performed every 8,000 steps, and the validation-based early stopping was applied when no progress was achieved in 3 consecutive validations.
The total number of 250 trials was performed for each architecture. Intermediate results of each trial were monitored and used to ensure only the top 10\% trials were allowed to continue. Details of the hyperparameter optimization are presented in Appendix~A.

\subsection{Results on WikiReading}
Although the main focus of our evaluation is the \WRR{} dataset; we additionally evaluate whether the Vanilla Dual-Source Transformer can improve the state-of-the-art on \WR{}.

\begin{table}
\renewcommand{\arraystretch}{0.9}
\centering
\begin{tabular}{lr}
\toprule
Model & \MF\\ \midrule
Basic s2s~\cite{hewlett-etal-2016-wikireading} &~70.8 \\
Placeholder s2s~\cite{choi-etal-2017-coarse} &~75.6 \\
\vspace{5pt}
SWEAR~\cite{hewlett-etal-2017-accurate} &~76.8   \\
Basic s2s (our run) &~74.8 \\
Vanilla Transformer              &          79.3  \\
Vanilla Dual-Source Transformer &  \textbf{82.4} \\
\bottomrule
\end{tabular}
\caption{Results on \WR{} (test set). \textit{Basic s2s} denotes the re-implemented model described in Section~\ref{sec:baseline}.}
\label{tab:2}

\end{table}

\begin{table*}
\centering
\renewcommand{\arraystretch}{0.9}

\begin{tabular}{lrrrrrrr}
\toprule
Model                    & unseen  & rare    & categorical & relational & exact match & long & test-B \\ \midrule
Basic seq2seq             & 2.0     & 30.2    &    84.9     &      50.2  &     71.1    &      56.4   &   75.2     \\
Vanilla Dual-Source       & 0.0     & 40.7    &    83.9     &      70.8  &     80.5    &      63.1   &   77.5      \\
Dual-Source RoBERTa       & 0.0     & 50.7    &    86.0     &      76.8  &     84.3    &      68.2   &   80.9      \\
Finetuned T5              & 10.9    & 53.8    &    86.3     &      73.4  &     83.4    &      65.9   &   80.3      \\ \bottomrule
\end{tabular}

\caption{Results on \WRR{} human-annotated test set supplemented with scores on diagnostics subsets. All scores are \MMF{}.}
\label{tab:diagnostics-results}
\end{table*}

We reproduced the \textit{Basic seq2seq} model. It achieved a~\MF score of $74.8$, which is 4 points higher than reported by~\citet{hewlett-etal-2016-wikireading}.
The difference may be caused by poor optimization in the original work.
Our dual-source solution achieves 82.4 and outperforms the previous state-of-the-art model by 5.6 \MF{} points. To measure the impact of using two encoders instead of one, we evaluated the Vanilla Single-source Transformer, which takes a~concatenated pair of article and property as its input. Our dual-source model outperformed its single-source counterpart by 3.1 points. Table \ref{tab:2} presents the final results.

\subsection{Results on WikiReading Recycled}

The results on \WR{} show that the Dual-Source Transformer is beneficial to the Property Extraction task. On \WRR{}, we supplement the evaluation with pretrained models: Dual-Source RoBERTa and T5.

Table~\ref{tab:diagnostics-results} presents \MMF{} scores on the annotated test set (test-B).
All the transformer-based models outperform the \textit{Basic seq2seq}.
The Dual-Source Transformer achieved $77.5$ \MMF{}. Its pretrained version, Dual-Source RoBERTa, improves the result by $1.4$ points. As the T5 model beats the Vanilla Dual-Source Transformer, we may conclude that even though the \WRR{} dataset is very large, the pretraining is crucial for this MPE task.
It is worth remembering that the results on \WR{} and \WRR{} are not comparable due to the dissimilarities in metrics and datasets. We will elaborate on that in section~\ref{s:daa}.

\section{Discussion and Analysis} \label{s:daa}

The final scores of transformer-based models differ slightly on \WRR{}.
In order to get more insight, we analyze the models on diagnostic sets described in Section~\ref{sec:diagnostics}. 

\paragraph{Impact of Property Frequency.} We provide two diagnostic sets related to property frequency: \textit{unseen} and \textit{rare}.
Both dual-source models failed on the \textit{unseen} subset.
These models ignored the \textit{unseen} properties from the input and did not generate any answer.
The best result was achieved by the T5 model ($10.9$ points), albeit it still does not meet expectations.

\begin{figure}[h]
    \centering
    \includegraphics[width=1.0\linewidth]{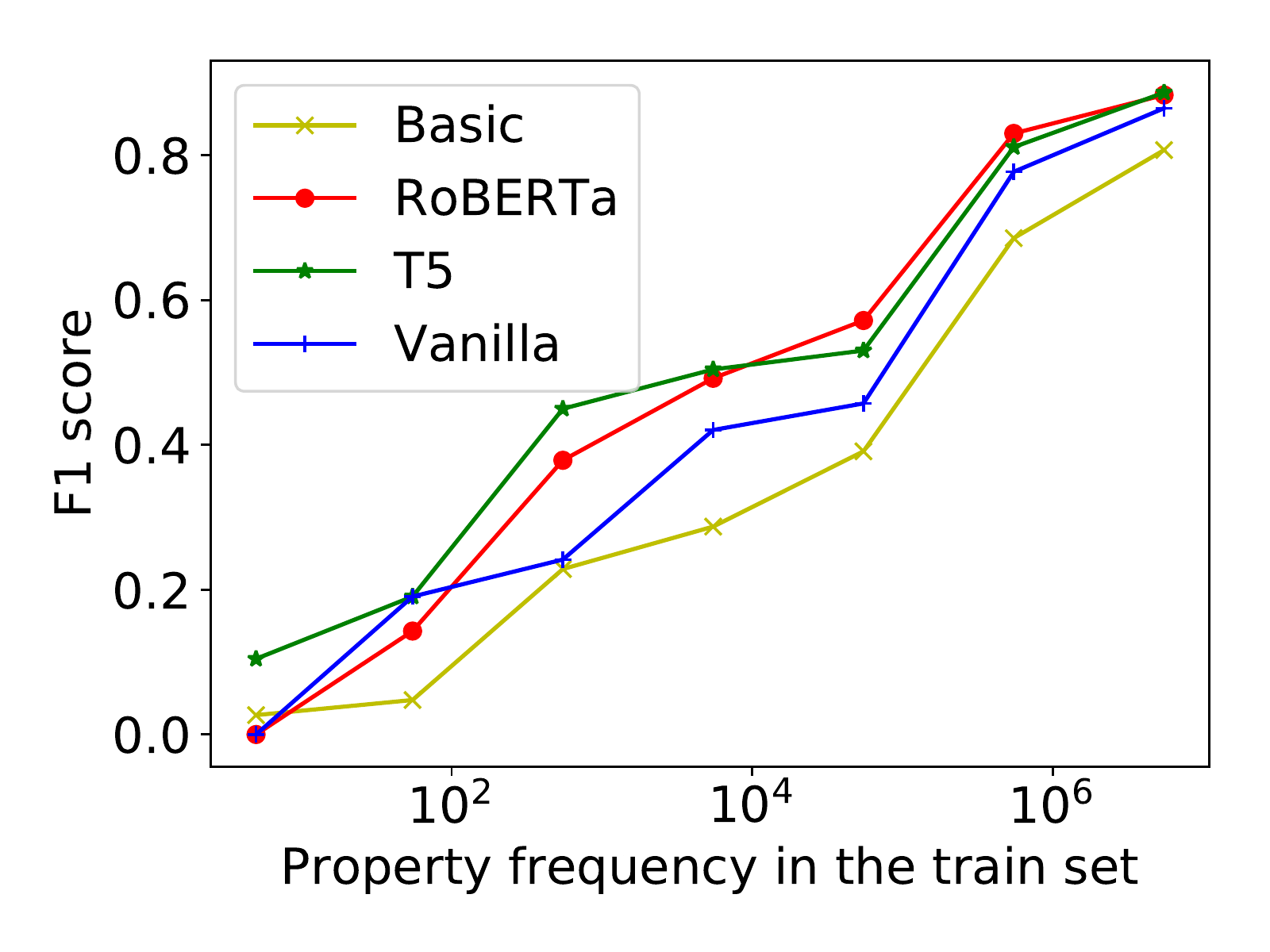}
    \caption{The relation of property frequency and \MMF{}. Both RoBERTa and Vanilla refer to Dual-Source Transformers.}\label{fig:freq}
\end{figure}

The results on the \textit{rare} subset show that the pretraining makes a~difference if properties are infrequent in the train set (Figure~\ref{fig:freq}).

\begin{figure}[h]
    \centering
    \includegraphics[width=1.0\linewidth]{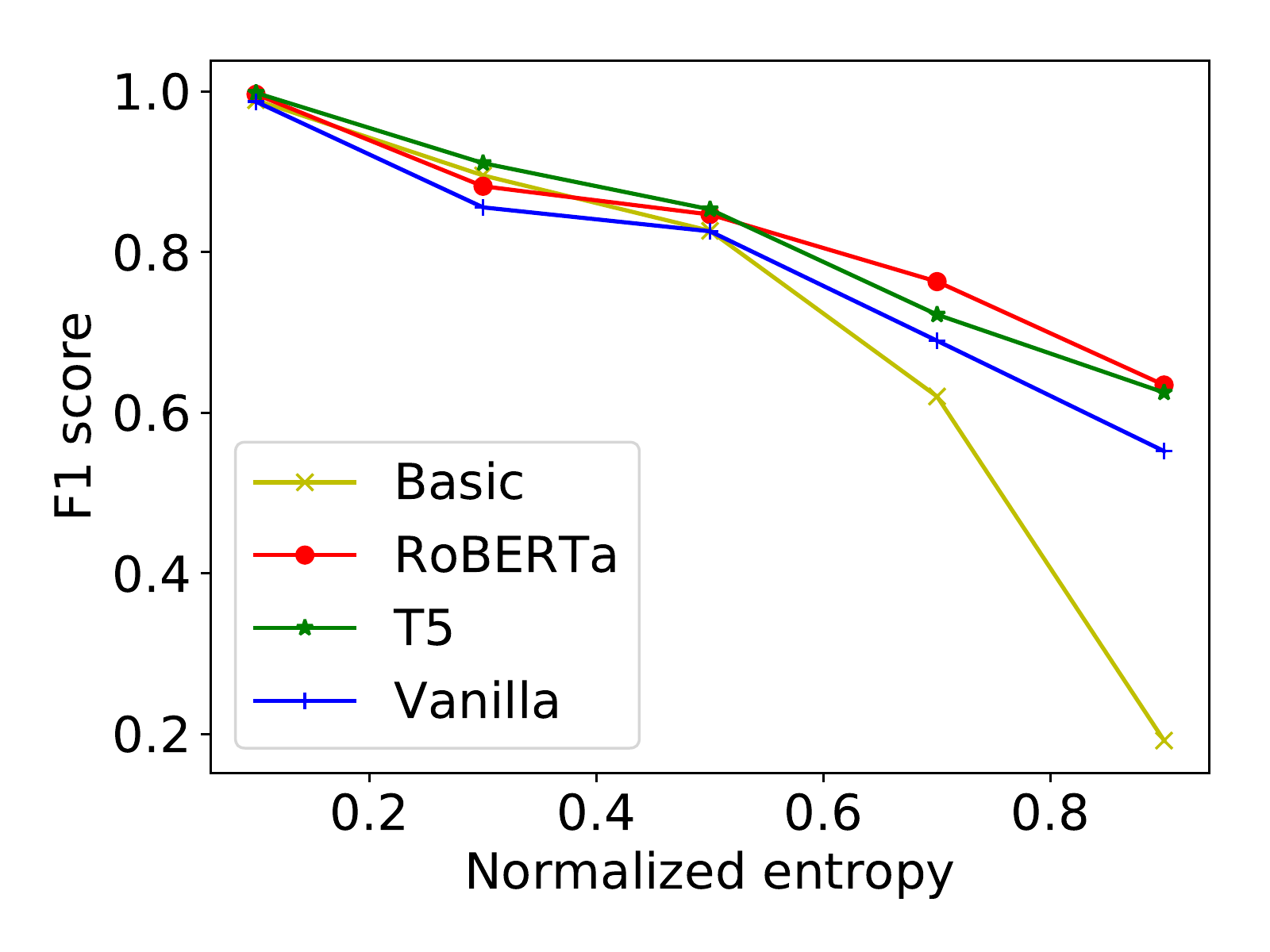}
    \caption{The relation of property normalized entropy and \MMF{}. Both RoBERTa and Vanilla refer to Dual-Source Transformers.}\label{fig:ce}
\end{figure}

\paragraph{Impact of Property Type.}
The extraction of some properties may be treated as a~classification task since the set of their valid values is limited. In this case, all models perform similarly and achieve approximately $85$ \MMF{}. The difficulty of the task increases proportionally to the normalized entropy value, which may lead to the divergence of model performances. This phenomenon is visible in the case of our Basic seq2seq, where the weakness is evident above the $0.5$ threshold.
The details are presented in Figure~\ref{fig:ce}.

\paragraph{Exact Match and Long Articles.}
The results from the exact match and long articles subsets are correlated with the scores attained on the test-B set; however, the absolute values achieved differ substantially. This is because the long article subset is more challenging, as the chance of an answer appearing in the constant-length prefix decreases with the length of the article. The use of recently introduced models like LongFormer~\cite{beltagy2020longformer} and BigBird~\cite{zaheer2020big} might decrease the gap in scores between long and average-length articles. On the other hand, system performance should increase when the answer is provided directly in the text, as can be found in the exact match subset.

\paragraph{Difficulty of Test Sets.}
To compare the difficulty of the \WR{} and \WRR{} test sets, we converted the outputs from the non-annotated \WRR{} test set (test-A) to \WR{} format, and calculated the \MF{}. With the Vanilla Dual-Source Transformer, we obtained $54.0$ \MF{}, $28.4$ points less than on \WR{}. This considerable decrease in score shows that the \WRR{} test-A set is more difficult than \WR{}. The reason behind this is that we removed leakage of articles between splits, and we also added more infrequent properties that are harder to answer.

\paragraph{Impact of Human Annotation.}
The Vanilla Dual-Source Transformer was evaluated on both \WRR{} test sets. It obtained \MMF{} of $62.6$ on the non annotated test-A set, while achieving $77.5$ on the annotated test-B. This discrepancy suggests that the linguists indeed succeeded to remove non-inferable properties. We anticipate that cleaning the train set in a~similar fashion could improve the stability of the training and the overall results.

\section{Summary}

We introduced \WRR{}---the first \MPE{} dataset with a~human-annotated test set. We provided strong baselines that improved the current state-of-the-art on \WR{} by a~large margin. The best-performing architecture was successfully adapted from Automatic Post-Editing systems.
We show that using pretrained language models increases the performance on the \WRR{} dataset significantly, despite its large size.

Additionally, we created diagnostic subsets to qualitatively assess model performance. The results on a~challenging subset of \textit{unseen} properties reveal that despite high overall scores, the evaluated systems fail to provide satisfactory performance. Low scores indicate an opportunity to improve, as these properties were verified by annotators and are expected to be answerable. We look forward to seeing models closing this gap and leading to remarkable progress in Machine Reading Comprehension.

The dataset and models, as well as their detailed configurations required for reproducibility, are publicly available.

\section*{Acknowledgements}

The Smart Growth Operational Programme supported this research under project no. POIR.01.01.01-00-0877/19 (A universal platform for robotic automation of processes requiring text comprehension, with a unique level of implementation and service automation).

\bibliography{ms}
\bibliographystyle{acl_natbib}
\appendix
\begin{table*}
\begin{tabularx}{\linewidth}{Xlccc}
\toprule
Parameter & Search space & Vanilla Dual-source & RoBERTa \\
\midrule
batch size & $2^{\{6, 7, 8, 9\}}$ & $2^9$ & $2^9$ \\
learning rate & $\expnumber{1}{-5}$, $\expnumber{5}{-5}$,.., $\expnumber{1}{-2}$ & $\expnumber{5}{-4}$ & $\expnumber{5}{-5}$ \\
lr scheduler & inverse sqrt, linear decay & linear & linear & \vspace{0.5em}\\
hidden dropout & \rdelim\}{4}{0mm}[ $0, 0.1$] & $0$ & $0.1$ \\
attention dropout & & $0$ & $0.1$ \\
activation dropout & & $0$ & $0$ \\
weight decay & & $0$ & $0.1$ & \vspace{0.5em}\\
encoder layers & \multirow{2}{*}{$1,..,6$} & $2$ & -- \\
decoder layers & & $2$ & $6$ & \vspace{0.5em}\\

embedding dim$^{*}$ & $2^{\{5, 6, .., 9\}}$ & $2^9$ & -- \\
ffn embedding dim$^{*}$ & $2^{\{6, 7, .., 11\}}$ & $2^7$ & -- \\
attention heads$^{*}$ & $2^{\{2, 3, 4, 5\}}$ & $2^3$ & -- \\
activation function$^{*}$ & ReLU, GELU & ReLU & GELU \\
learned positional emb$^{*}$ & true, false & false & -- \\
share all emb & true, false & false & -- \\
\bottomrule
\end{tabularx}
\caption{Search space considered and hyperparameters determined as optimal when the validation set of WRR is considered. The $^{*}$ symbol denotes tied hyperparameters set to the same values for both encoder and decoder where applicable. The use of pretrained RoBERTa model resulted in the necessity to stick with several architectural choices signalized by -- character.\label{tab:hparam}}
\end{table*}

\section{Hyperparameter Search\label{sec:appendix}}

Table~\ref{tab:hparam} summarizes search space considered and hyperparameters determined as optimal when the validation set of WRR is considered. 

Hyperparameters for WRR were optimized using the Tree-structured Parzen Estimator with additional heuristics and Gaussian priors resulting from the default settings proposed for this sampler in the Optuna framework.
An evaluation was performed every 8,000 steps, and the validation-based early stopping was applied when no progress was achieved in three consecutive validations.
Intermediate results of each trial (results from every validation) were monitored and used to stop unpromising training earlier.

The trial was pruned in the case its best intermediate value was in the bottom 90 percentiles among trials at the same step (only the top 10\% of trials were allowed to continue the training). This process was disabled until five trials finished.

The total number of 250 trials was performed for each architecture.

\section{Basic seq2seq Replication Details}

Since the basic seq2seq model description missed some essential details, they had to be assumed before model training. For example, we supposed that the model consisted of unidirectional LSTMs. It was trained with mean (per word) cross-entropy loss until no progress was observed for 10 consecutive validations occurring every 10,000 updates. Input and output sequences were tokenized and lowercased. Besides, 
and truecasing was applied to the output. We use syntok\footnote{\url{https://github.com/fnl/syntok}} tokenizer and a~simple RNN-based truecaser proposed by~\citet{susanto-etal-2016-learning}. During inference, we used a~beam size of 8.
The rest of the parameters followed the description provided by the authors.

\end{document}